\documentclass[runningheads]{llncs}
\usepackage{graphicx}
\usepackage{textcomp}
\usepackage{comment}
\usepackage{todonotes}
\usepackage{color}
\usepackage{array}
\newcolumntype{x}[1]{>{\centering\arraybackslash\hspace{0pt}}p{#1}}

\usepackage{multirow}

\begin{document}
\title{Fairness in Cardiac MR Image Analysis: An Investigation of Bias Due to Data Imbalance in Deep Learning Based Segmentation}

\author{Esther Puyol-Ant\'on \inst{1} \and 
Bram Ruijsink \inst{1,2} \and
Stefan K. Piechnik \inst{7} \and
Stefan Neubauer \inst{7} \and
Steffen E. Petersen \inst{3,4,5,6} \and
Reza Razavi \inst{1,2} \and
Andrew P. King \inst{1}}
\authorrunning{E Puyol-Ant\'on et al.}   
\institute{School of Biomedical Engineering \& Imaging Sciences, King\textquotesingle s College London, UK \and Guy’s and St Thomas\textquotesingle{} Hospital, London, UK. \and William Harvey Research Institute, NIHR Barts Biomedical Research Centre, Queen Mary University London, Charterhouse Square, London, EC1M 6BQ, UK \and Barts Heart Centre, St Bartholomew’s Hospital, Barts Health NHS Trust, West Smithfield, EC1A 7BE, London, UK \and Health Data Research UK, London, UK \and Alan Turing Institute, London, UK \and Division of Cardiovascular Medicine, Radcliffe Department of Medicine, University of Oxford, UK.}


\maketitle              

\begin{abstract} 
The subject of ‘fairness’ in artificial intelligence (AI) refers to assessing AI algorithms for potential bias based on demographic characteristics such as race and gender, and the development of algorithms to address this bias. Most applications to date have been in computer vision, although some work in healthcare has started to emerge. The use of deep learning (DL) in cardiac MR segmentation has led to impressive results in recent years, and such techniques are starting to be translated into clinical practice. However, no work has yet investigated the fairness of such models. In this work, we perform such an analysis for racial/gender groups, focusing on the problem of training data imbalance, using a nnU-Net model trained and evaluated on cine short axis cardiac MR data from the UK Biobank dataset, consisting of 5,903 subjects from 6 different racial groups. We find statistically significant differences in Dice performance between different racial groups. To reduce the racial bias, we investigated three strategies: (1) stratified batch sampling, in which batch sampling is stratified to ensure balance between racial groups; (2) fair meta-learning for segmentation, in which a DL classifier is trained to classify race and jointly optimized with the segmentation model; and (3) protected group models, in which a different segmentation model is trained for each racial group. We also compared the results to the scenario where we have a perfectly balanced database. To assess fairness we used the standard deviation (SD) and skewed error ratio (SER) of the average Dice values. Our results demonstrate that the racial bias results from the use of imbalanced training data, and that all proposed bias mitigation strategies improved fairness, with the best SD and SER resulting from the use of protected group models.
\end{abstract}

\section{Introduction}
Fairness in artificial intelligence (AI) is a relatively new but fast-growing research field which deals with assessing and addressing potential bias in AI models. For example, an early landmark paper \cite{Buolamwini2018} found differences in performance of a video-based gender classification model for different racial groups. With AI models starting to be deployed in the real world it is essential that the benefits of AI are shared equitably according to race, gender and other demographic characteristics, and so efforts to ensure the fairness of deployed models have generated much interest. Most work so far has focused on computer vision problems  but some applications in healthcare are starting to emerge \cite{obermeyer2019dissecting,wilder2020clinical}.

Recently, deep learning (DL) models have shown remarkable success in automating many medical image segmentation tasks. In cardiology, human-level performance in segmenting the main structures of the heart has been reported \cite{Bernard2018}, and researchers have proposed to use these models for tasks such as automating cardiac functional quantification \cite{Ruijsink2020}. These methods are now starting to move towards wider clinical translation.

It has long been well understood that cardiac structure and function, as well as the mechanisms leading to cardiovascular disease, vary according to demographic characteristics such as race and gender \cite{kishi2015race}. For example, the Multi-Ethnic Study of Atherosclerosis (MESA) \cite{yoneyama2017cardiovascular} showed that there are racial differences in regional left ventricular (LV) systolic function in a large cohort study of adults. In addition, there are profound race-associated disparities among those who are affected by and die from cardiovascular disease \cite{mody2012most}. Inequalities in detection of disease likely play a role in these differences. Therefore, it is surprising that no work to date has investigated potential bias in AI models for cardiac image analysis. To the best of our knowledge, the closest related work was on assessing differences in radiomics features by gender \cite{raisi2020variation}.

In this paper, we perform the first analysis of the fairness of DL-based cardiac MR segmentation  models, focusing in particular on the impact of gender or race imbalance in the training data. To the best of our knowledge, this is also the first analysis of the fairness of segmentation models in general. We discover significant bias in performance between racial groups when trained using one of the largest and widely used public databases of cardiac magnetic resonance (MR) data (the UK Biobank). We propose algorithms to mitigate this bias, resulting in a fairer segmentation model that ensures that no racial group will be disadvantaged when segmentations of their cardiac MR data are used to inform clinical management.

\section{Background}
In recent years, addressing fairness concerns in AI models has been an active research area. To date, most previous works have focused on classification tasks such as image recognition, although Hwang \textit{et al.} \cite{hwang2020fairfacegan} investigated bias in image-to-image translation. In an extensive search, we did not find any previous work on bias mitigation for image segmentation tasks.

At a high level, fairness techniques can be grouped into two categories: (i) `fairness through awareness', also known as bias mitigation strategies or discrimination-aware classification, that aim to make the AI algorithms more aware of the protected attributes\footnote{In fair AI, the \emph{protected attribute(s)} are the ones for which fairness needs to be ensured, e.g. gender or race. A set of samples with the same value(s) for the protected attribute(s) are known as a \emph{protected group}.} by making predictions independently for each protected group \cite{dwork2012fairness}; and (ii) `fairness through unawareness', which assumes that if the model is unaware of the protected attributes while making decisions, the decisions will be fair. This second approach has been the most common one used to train AI models. However, it has been shown to be unsuccessful in many cases \cite{wang2020towards,zhang2018mitigating} due to the correlation of protected attributes with other variables in the data.

Focusing now on bias mitigation algorithms (a.k.a. `fairness through awareness'), DL pipelines contain three possible points of intervention to mitigate unwanted bias: the training data, the learning procedure, and the output predictions, and these are associated with three corresponding classes of bias mitigation strategy: pre-processing, in-processing, and post-processing:\\
\textbf{1) Pre-processing approaches} modify the training dataset to remove the discrimination before training an AI model. Common strategies used are under-sampling, over-sampling \cite{wang2020towards} or sample weighting \cite{kamiran2012data} to neutralize discriminatory effects; data generation \cite{ngxande2020bias} or data augmentation \cite{lu2020gender} using generative adversarial networks to balance the training dataset; or training using a balanced dataset \cite{wang2020mitigating}. \\
\textbf{2) In-processing approaches} try to modify state-of-the-art learning algorithms in order to remove discrimination during the model training process via model regularization. There are two main approaches: implicit regularization that adds implicit constraints which disentangle the association between model prediction and fairness sensitive attributes \cite{das2018mitigating}; and explicit regularization that adds explicit constraints through updating the model's loss function to minimize the performance difference between different protected groups \cite{ngxande2020bias,xu2020investigating}. \\
\textbf{3) Post-processing approaches} correct the output of an existing algorithm to satisfy the fairness requirements. Common strategies used are equalized odds post-processing \cite{hardt2016equality} that solves a linear program to find probabilities with which to change output labels to optimize equalized odds; calibrated equalized odds post-processing \cite{pleiss2017fairness} that optimizes over calibrated classifier score outputs to find probabilities with which to change output labels with an equalized odds objective; and reject option classification \cite{kamiran2012decision} that gives favorable outcomes to unprivileged protected groups and unfavorable outcomes to privileged protected groups in a confidence band around the decision boundary with the highest uncertainty.

\section{Methods}
\label{sec:methods}
\begin{figure}[ht]%
    \centering
    \includegraphics[width=\textwidth]{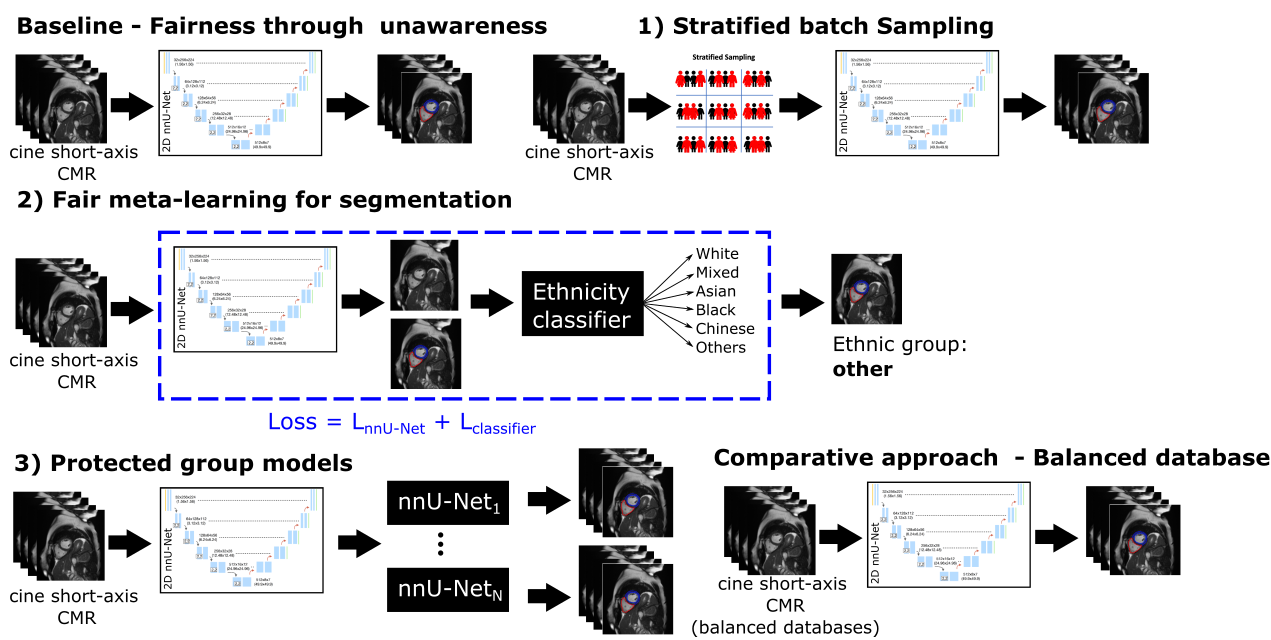}
    \caption{Diagram showing the different strategies used for bias mitigation. Baseline is based on fairness through  unawareness, and acts as the baseline approach. (1)-(3) are three approaches based on fairness through  awareness: (1) stratified batch sampling, (2) `fair meta-learning for segmentation' and (3) protected group models. The comparative approach of using a smaller balanced database allows us to test the hypothesis that observed bias is due to data imbalance.}
    \label{fig:overview}%
\end{figure}

To investigate whether training data imbalance can lead to bias between racial and/or gender groups in automated cardiac MR segmentation, we conducted a comparative study using five different approaches. The baseline approach was based on fairness through  unawareness. To enable us to test the hypothesis that any observed bias was due to data imbalance, we also trained the segmentation network using a (smaller) race and gender balanced database. We subsequently investigated three approaches based on fairness through  awareness (two pre-processing approaches: stratified batch sampling and protected group models and one in-processing method that we call a `fair meta-learning for segmentation'). 
These approaches are illustrated in detail in Fig. \ref{fig:overview}.

\noindent
\textbf{Segmentation network}: For all of the approaches we used the nnU-Net network \cite{isensee2021nnu} for automatic segmentation of the left ventricle blood pool (LVBP), left ventricular myocardium (LVM) and right ventricle blood pool (RVBP) from cine short-axis cardiac MR slices at end diastole (ED) and end systole (ES). Details of the data used and experimental set up are provided in Section \ref{sec:results}.\\
\textbf{Baseline - Fairness through  unawareness:} We trained the nnU-net model blinded to the protected attributes (i.e. race and gender). This will be considered the baseline approach in the rest of the paper and its performance can be directly compared with other published cardiac MR segmentation techniques \cite{Ruijsink2020,Bernard2018}.\\
\textbf{Comparative approach - Balanced database:} This strategy aims to evaluate the scenario where the database is perfectly balanced. To this end, we randomly selected for each racial group the same number of subjects (N=87), which corresponds to the number of subjects in the smallest racial group. In total, 522 subjects were used for training this model. \\
\textbf{Approach 1 - Stratified batch sampling:} This strategy aims to modify the training sampling strategy to remove the discrimination before training. For each training batch, the data are stratified by the protected attribute(s) and samples are selected to ensure that each protected group is equally represented.
This approach has previously been used to train fair classifiers \cite{kamiran2012data}.
\\
\textbf{Approach 2 - Fair meta-learning for segmentation:} This strategy aims to add a meta-fair classifier to the segmentation network to train a model that not only segments cardiac MR images but also performs classification of the protected attribute(s). As the classifier we used a DenseNet network \cite{huang2017densely} and the input of the classifier was the cardiac MR image as one channel and the output of the nnU-Net segmentation network as a second channel. We formulate the problem as a multi-task learning problem where both networks are jointly optimized. The idea is that the classification network prevents the learning of a dominant group from negatively impacting the learning of another one. Both networks were first trained for 500 epochs independently, and then jointly trained for another 500 epochs without deep supervision. This is a novel approach for segmentation fairness but is based on the work proposed by Xu \textit{et al.,} \cite{xu2020investigating} for classification fairness, which combined a ResNet-18 network for facial expression recognition with a fully connected network for protected attribute classification.\\
\textbf{Approach 3 - Protected group models:} In contrast to Approaches 1 and 2, this strategy assumes that the protected attributes are available at inference time (as well as training time) and aims to train a different segmentation model for each protected group.  As the number of subjects for each protected group can vary, which could impact performance, we initially trained the nnU-Net model for 500 epochs using the full training database, and then fine-tuned independent nnU-Net models for each of the protected groups for another 500 epochs. This approach is similar in concept to \cite{wang2020mitigating} in which classifiers were first trained using an unbalanced database and then fine-tuned using balanced ones. However, we choose to fine-tune using only individual protected group data and we are not aware of previous work that has taken this approach. \\
\textbf{Evaluation metrics:}
In the literature, several fairness metrics have been proposed but because fairness research has so far focused on classification tasks all of them are tailored for such tasks. Because we deal with a segmentation task, we propose to evaluate our approaches using a segmentation overlap metric and two new segmentation fairness metrics. To measure segmentation overlap we use the average Dice similarity coefficient (DSC) over the three classes, i.e. LVBP, LVM and RVBP. As fairness metrics, we utilize the standard deviation (SD) and skewed error ratio (SER) of the average DSC values. The standard deviation reflects the amount of dispersion of the average DSC values between different protected groups. The SER is computed by the ratio of the highest error rate to the lowest error rate among different protected groups and it can be formulated as $SER = \frac{\max_g (1-DSC_g)}{\min_g (1-DSC_g)}$, where $g$ are the protected groups. The SD and SER fairness metrics were adapted from \cite{wang2020mitigating}, which used classification accuracy and classification error rate for SD and SER respectively.

\section{Materials and Experiments}
\label{sec:results} 
\begin{figure}[ht]%
    \centering
    \includegraphics[width=0.6\textwidth]{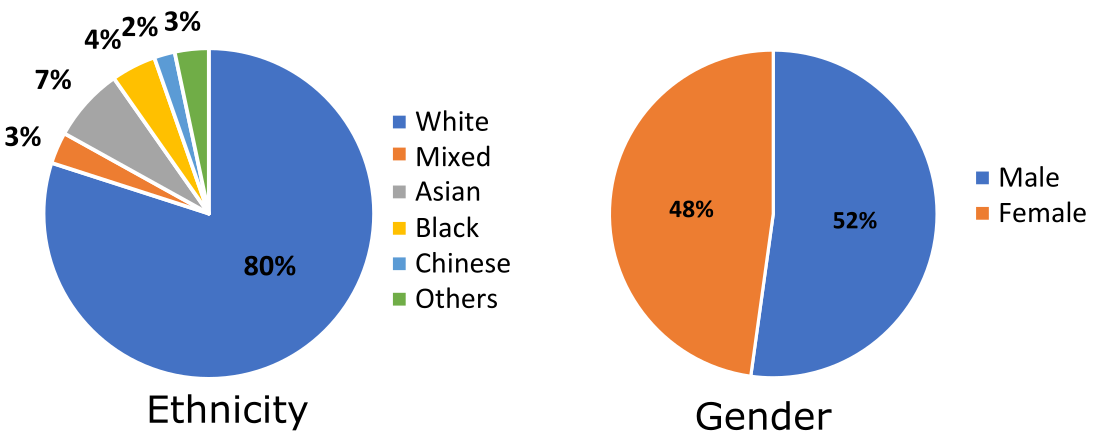}
    \caption{The percentage of gender and different races in the UK Biobank dataset used in this paper.}
    \label{fig:gender_race_distribution}
\end{figure}

We demonstrated our approach for fairness in DL-based cardiac MR segmentation using data from the UK Biobank \cite{petersen2015uk}. The dataset used consisted of ED and ES short-axis cine cardiac MR images of 5,903 subjects (61.5 $\pm$ 7.1 years). Data on race and gender were obtained from the UK Biobank database and their distribution is summarized in Figure \ref{fig:gender_race_distribution}. For all subjects, the LV endocardial and epicardial borders and the RV endocardial border were manually traced at ED and ES frames using the cvi42 software (version 5.1.1, Circle Cardiovascular Imaging Inc., Calgary, Alberta, Canada). Each segmentation was generated by 1 of 10 experts who followed the same guidelines and were blinded to race and gender. Each expert contoured a random sample of images containing different races and genders. For all approaches (apart from the balanced database comparative approach), we used the same random split into training/validation/test sets of 4,723/590/590 subjects respectively. For the balanced database approach the training/validation/test set sizes were 417, 105 and 590. All models were trained on a NVIDIA GeForce GTX TITAN X. All networks were trained for 1,000 epochs and optimized using stochastic gradient descent with ‘poly’ learning rate policy (initial learning rate of 0.01 and Nesterov momentum of 0.99).

\noindent
\textbf{Bias assessment:} The first experiment aims to assess gender and racial bias for the baseline approach (i.e. `fairness through  unawareness'). Table \ref{table:1} shows the DSC values for LVBP, LVM and RVBP at ED and ES as well as the overall average DSC. Unpaired Student’s $t-$tests (significant difference reported for $p-$value $<$ 0.01 with Scheffe post hoc test) were used for comparison between the full test set and each of the protected groups for the average DSC. From the results, we can make several important observations. First, there is a clear racial bias. For example, when the racial distribution is 80\% white and 20\% others, the nnU-net model reaches 93.51\% average DSC for white subjects, but this dramatically decreases to less than 86\% for black subjects and less than 85\% for mixed race subjects. Second, the results show that there is no significant gender bias. This suggests that the accuracy of the baseline segmentation model for each group is correlated with its representation in the training set. The results of the comparative approach (Balanced database, see bottom row of Table \ref{table:2}) support this conclusion, achieving the lowest SD and SER. As expected, the smaller training database size led to a significant reduction in accuracy for all racial groups.

\noindent
\textbf{Bias mitigation strategies:}
Since Table \ref{table:1} showed that the baseline segmentation model had a racial bias but no gender bias, in this second experiment we focus only on the use of bias mitigation techniques to reduce racial bias. Table \ref{table:2} shows the comparison between the baseline approach and the three approaches for bias mitigation. The results show that all mitigation strategies reduce the racial bias but that Approach 3 (Protected group models) achieved the lowest SD and SER. Note, however, that Approach 3 requires knowledge of the protected attributes at inference time. The best-performing approach that does not require such knowledge is Approach 1 (Stratified batch sampling). For Approach 2 (Fair meta-learning for segmentation), the average accuracy of the protected attribute classifier was 0.77 and the precision and recall values per group varied between 0.69 and 0.89.

\begin{table}[ht] 
\caption{Dice similarity coefficient (DSC) for the LV blood pool (LVBP), LV myocardium (LVM) and RV blood pool (RVBP) at end diastole (ED) and end systole (ES) for Baseline - fairness through  unawareness.  The `Avg' column that represents the average DSC values across LVBP, LVM and RVBP at both ED and ES has been boldfaced for easier comparison with Table \ref{table:2}. The first row reports the DSC for the full database,  the second and third rows report DSC by gender and the remaining rows report DSC by racial group. Asterisks indicate statistically significant differences between the full population and each protected group for the average DSC.}  
\centering
\begin{tabular}{p{2cm} x{0.1cm} x{1.5cm}x{1.5cm}x{1.5cm} x{0.1cm} x{1.5cm}x{1.5cm}x{1.5cm} x{0.1cm} x{1.5cm}|}
& & \multicolumn{9}{c}{\textbf{DSC (\%) for Baseline \textemdash Fairness through  unawareness}} \\ \cline{2-11}
& \multicolumn{1}{|c}{} & \multicolumn{3}{c}{\textbf{ED}} & \multicolumn{1}{|c}{} & \multicolumn{3}{c}{\textbf{ES}} & \multicolumn{1}{|c}{} & \multirow{2}{*}{\textbf{Avg}}\\ 
& \multicolumn{1}{|c}{} & LVBP  & LVM   & RVBP  & \multicolumn{1}{|c}{} & LVBP  & LVM   & RVBP & \multicolumn{1}{|c}{} & \\ \hline
\multicolumn{1}{|c}{Total} & \multicolumn{1}{|c}{} & 93.48 & 83.12 & 89.37 & \multicolumn{1}{|c}{} & 89.37 & 86.31 & 80.61 & \multicolumn{1}{|c}{} & \textbf{87.05}\\ \hline \hline
\multicolumn{1}{|c}{Male} & \multicolumn{1}{|c}{} & 93.58 & 83.51 & 88.82 & \multicolumn{1}{|c}{} & 90.68 & 85.31 & 81.00 & \multicolumn{1}{|c}{} & \textbf{87.02}\\
\multicolumn{1}{|c}{Female}  & \multicolumn{1}{|c}{} & 93.39 & 82.71 & 89.90 & \multicolumn{1}{|c}{} & 89.59 & 86.60 & 80.21 & \multicolumn{1}{|c}{} & \textbf{87.07}\\ \hline
\multicolumn{1}{|c}{White} & \multicolumn{1}{|c}{} & 97.33 & 93.08 & 94.09 & \multicolumn{1}{|c}{} & 95.06 & 90.58 & 90.88 & \multicolumn{1}{|c}{} & \textbf{93.51*}\\
\multicolumn{1}{|c}{Mixed} & \multicolumn{1}{|c}{} & 92.70 & 78.94 & 86.91 & \multicolumn{1}{|c}{} & 86.70 & 82.54 & 79.32 & \multicolumn{1}{|c}{} & \textbf{84.52*}\\
\multicolumn{1}{|c}{Asian} & \multicolumn{1}{|c}{} & 94.53 & 87.33 & 90.51 & \multicolumn{1}{|c}{} & 90.13 & 88.94 & 81.94 & \multicolumn{1}{|c}{} & \textbf{88.90*}\\
\multicolumn{1}{|c}{Black}  & \multicolumn{1}{|c}{} & 92.77 & 85.93 & 89.49 & \multicolumn{1}{|c}{} & 89.42 & 85.74 & 71.91 & \multicolumn{1}{|c}{} & \textbf{85.88*}\\
\multicolumn{1}{|c}{Chinese} & \multicolumn{1}{|c}{} & 91.81 & 74.51 & 85.74 & \multicolumn{1}{|c}{} & 86.39 & 85.12 & 79.34 & \multicolumn{1}{|c}{} & \textbf{83.82*}\\
\multicolumn{1}{|c}{Others} & \multicolumn{1}{|c}{} & 91.74 & 78.94 & 89.50 & \multicolumn{1}{|c}{} & 88.53 & 84.96 & 80.27 & \multicolumn{1}{|c}{} & \textbf{85.66*}\\ \hline 
\end{tabular}
\label{table:1}
\end{table}

\begin{figure}[ht]%
    \centering
    \includegraphics[width=\textwidth]{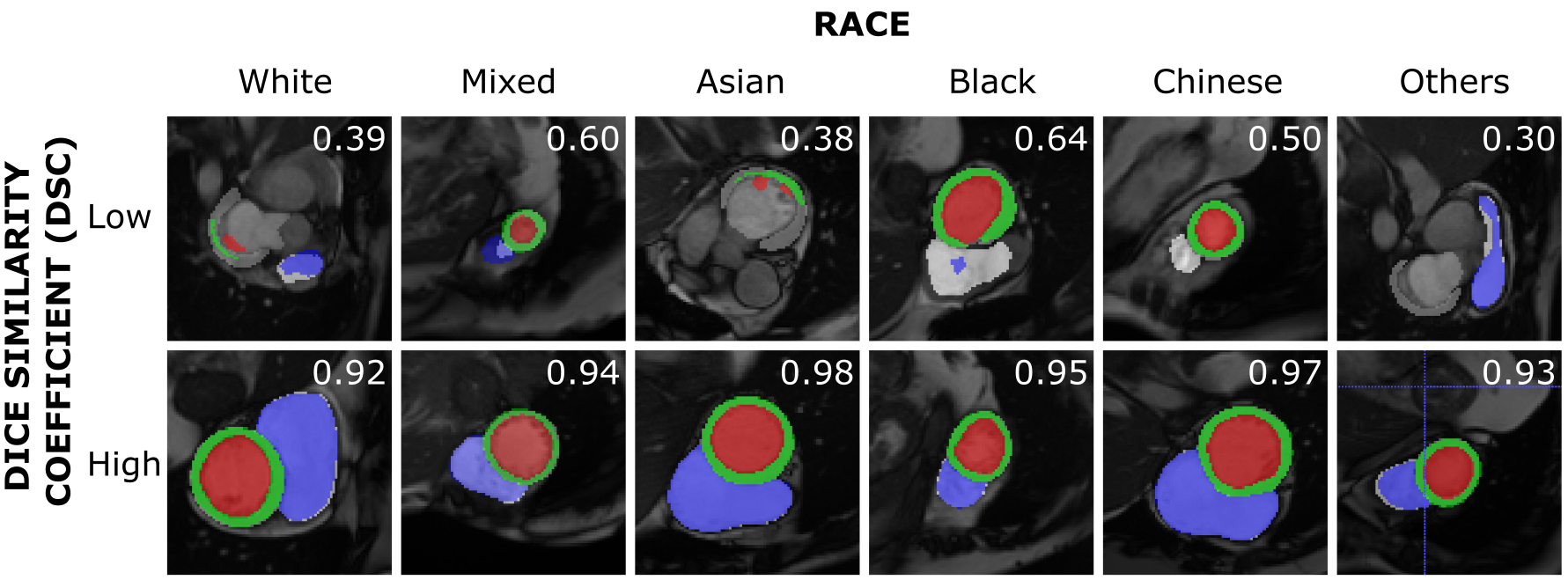}
    \caption{Illustration of the segmentation results at ED for each racial group with high (top row) and low (bottom row) DSC.}
    \label{fig:sample_cases}
\end{figure}

Figure \ref{fig:sample_cases} shows some sample segmentation results for different racial groups with both high and low DSC.

\begin{table}[ht] 
\caption{Comparison between baseline model (i.e. `fairness through  unawareness') and three approaches for bias mitigation. The values shown for each race represent the average DSC values across LVBP, LVM and RVBP at both ED and ES. The `Avg' column is the average such value across all races (without weighting for the number of subjects in each group). We boldface the SD (lower is better) and SER (1 is the best) metrics since these are the important fairness criteria. The bottom row shows the results for using a smaller gender and race balanced training database.}
\centering
\begin{tabular}{l| ccccccc |cc}
\hline
\multirow{2}{*}{\textbf{Approach}} & \multicolumn{7}{c|}{\textbf{Segmentation}} & \multicolumn{2}{c}{\textbf{Fairness}} \\ \cline{2-10}
& White & Mixed & Asian & Black & Chinese & Others & Avg & SD & SER\\ \hline
Baseline - Fairness through  unawareness & 93.51 & 84.52 & 88.90 & 85.88 & 87.63 & 85.66 & 87.68 & \textbf{3.25} & \textbf{2.38} \\
1. Stratified batch sampling & 90.88 & 93.84 & 93.65 & 93.07 & 94.35 & 93.50 & 93.22 & \textbf{1.22} & \textbf{1.62} \\
2. Fair meta-learning for segmentation & 92.75 & 88.03 & 90.64 & 89.60 & 88.18 & 88.27 & 89.58 & \textbf{1.86} & \textbf{1.65} \\
3. Protected group models & 91.03 & 93.17 & 93.34 & 92.15 & 93.04 & 93.08 & 92.64 & \textbf{0.89} & \textbf{1.35}
\\
Comparative approach - Balanced database &   79.32 & 80.98 & 80.37 & 79.78 & 80.82 & 80.72 & 80.33 & \textbf{0.65} & \textbf{1.09}\\ \hline
\end{tabular}
\label{table:2}
\end{table}

\section{Discussion}
\label{sec:discussion}
To the best of our knowledge, this paper has presented the first study of fairness in AI-based image segmentation, focusing on the problem of data imbalance. We have shown, for the first time, that racial bias exists in DL-based cardiac MR segmentation models. Our hypothesis is that this bias is a result of the unbalanced nature of the training data, and this is supported by the results which show that there is racial bias but not gender bias when trained using the UK Biobank database, which is gender-balanced but not race-balanced. We also found very low bias when training using a balanced database. One reason why bias has yet to be investigated in DL-based cardiac MR segmentation is that in current clinical practice the outputs of such models are typically  manually modified by a clinician. However, this modification process is time-consuming and prone to error, so does not eliminate the bias completely. Furthermore, in the future, as AI models become more clinically accepted, it may be that the role of clinician oversight decreases or is removed completely, in which case lack of bias in DL-based segmentation models will become a critical feature.

We have proposed three `fairness through awareness' bias mitigation strategies. The first strategy (Stratified batch sampling) ensures that in each batch all protected groups are equally represented. The second strategy (Fair meta-learning for segmentation) aims to make the model more aware of the protected attribute(s) by simultaneously training a protected attribute classifier. The last strategy (Protected group models) aims to train independent segmentation models for each protected group. Our results show that all three strategies improve fairness, with lower SD and SER values compared to the baseline model. These results are in keeping with previous works where the strategy of `fairness through unawareness' has shown to be insufficient in preventing bias in several problems \cite{wang2020towards,zhang2018mitigating}.

It is important to note that the best-performing bias mitigation strategy (Approach 3) requires knowledge of the protected attribute at inference time. In general in medical image segmentation, such patient information is available via electronic health records. However, there are likely to be scenarios in which it is not available. In the case where the protected attribute(s) is not available at inference time it would be possible to use either Approach 1 or Approach 2. Approach 3 could be seen as a `best achievable' model for such situations or where the protected attributes have been inadequately registered.

In conclusion, in this work we have highlighted the concerning issue of bias in DL-based image segmentation models. We have proposed three bias mitigation techniques, which were inspired by work from the literature on fairness in classification but which are all novel in the context of image segmentation. In the future, we aim to extend the current work to other kinds of existing fairness interventions, such as combinations of pre-processing, in-processing and post-processing strategies, as well as extending the analysis to other imaging modalities.

\section*{Acknowledgements}
This work was supported by the EPSRC (EP/R005516/1 and EP/P001009/1), the Wellcome EPSRC Centre for Medical Engineering at the School of Biomedical Engineering and Imaging Sciences, King\textquotesingle s College London (WT 203148/Z/16/Z) and has been conducted using the UK Biobank Resource under application numbers 17806 and 2964. SEP, SN and SKP acknowledge the BHF for funding the manual analysis to create a cardiac MR imaging reference standard for the UK Biobank imaging resource in 5000 CMR scans (PG/14/89/31194). We also acknowledge the following funding sources: the NIHR Biomedical Research Centre at Barts, EU’s Horizon 2020 (grant no. 825903, euCanSHare project), the CAP-AI programme funded by the ERDF and Barts Charity, HDR UK, the Oxford NIHR Biomedical Research Centre, the Oxford BHF Centre of Research Excellence and the MRC eMedLab Medical Bioinformatics infrastructure (MR/L016311/1).

\bibliographystyle{splncs04}
\bibliography{refs}

\end{document}